\documentclass[10pt,twocolumn,letterpaper]{article}

\usepackage{times}
\usepackage{authblk}
\usepackage{epsfig}
\usepackage{graphicx}
\usepackage{amsmath}
\usepackage[font={small}]{caption}
\usepackage{amssymb}
\usepackage{physics}
\usepackage{algorithmic}
\usepackage[caption=false,font=small]{subfig}
\usepackage{algorithm}
\usepackage{pgfplots}
\usepackage{tikzinclude}
\usepackage{booktabs}
\usepackage{bm}
\usepackage{tikz}
\usetikzlibrary{positioning,shapes,arrows}
\usepackage[margin=2cm]{geometry}

\pgfplotsset{every axis/.append style={
                   label style={font=\footnotesize},
                   tick label style={font=\footnotesize}  
                    }}

\tikzstyle{block} = [rectangle, draw, fill=blue!20, 
    text width=8em, text centered, rounded corners, minimum height=3em]
\tikzstyle{out-block} = [rectangle, draw, fill=blue!40, 
    text width=15em, text centered, rounded corners, minimum height=3em]
\tikzstyle{back} = [rectangle, draw, fill=orange!20, 
    text width=\linewidth , rounded corners, minimum height=11em]
\tikzstyle{back2} = [rectangle, draw, fill=orange!20, 
    text width=\linewidth, rounded corners, minimum height=11em]
\tikzstyle{every node}=[font=\footnotesize]

\begin{document}

%%%%%%%%% TITLE
\title{Graph Cut Segmentation Methods Revisited with a Quantum Algorithm}

\author[1]{Lisa Tse}
\author[2]{Peter Mountney}
\author[3]{Paul Klein}
\author[4]{Simone Severini}
\affil[1,4]{Department of Computer Science, University College London, London, UK}
\affil[2,3]{Siemens Healthineers, Medical Imaging Technologies, Princeton, NJ, USA}
\maketitle

%%%%%%%%% ABSTRACT
\begin{abstract}
\textit{The design and performance of computer vision algorithms are greatly influenced by the hardware on which they are implemented. CPUs, multi-core CPUs, FPGAs and GPUs have inspired new algorithms and enabled existing ideas to be realized. This is notably the case with GPUs, which has significantly changed the landscape of computer vision research through deep learning. As the end of Moore’s law approaches, researchers and hardware manufacturers are exploring alternative hardware computing paradigms. Quantum computers are a very promising alternative and offer polynomial or even exponential speed-ups over conventional computing for some problems. This paper presents a novel approach to image segmentation that uses new quantum computing hardware. Segmentation is formulated as a graph cut problem that can be mapped to the quantum approximate optimization algorithm (QAOA). This algorithm can be implemented on current and near-term quantum computers. Encouraging results are presented on artificial and medical imaging data. This represents an important, practical step towards leveraging quantum computers for computer vision.}
\end{abstract}

\section{Introduction}
Advances in algorithms have driven the field of computer vision research forward and led to significant improvements in performance over the years. However, the nature of those algorithms is heavily influenced by the underlying hardware. A wide body of work, including seminal papers on edge detection \cite{pm1} and optical flow \cite{pm2} started with single core CPUs. The introduction of multi-core CPUs led to new real-time computer vision systems \cite{pm3}. More recently, advances in GPUs motivated researchers to revisit the idea of neural networks \cite{pm4}, leading to the widespread adoption of deep learning.

As we approach the end of Moore's Law \cite{pm5}, researchers and hardware manufacturers are actively investigating alternative computing paradigms. Alternative hardware includes visual processing units, neuromorphic, optical, biological and quantum computers \cite{neuro1, optical1972, biological_comp1, biological_comp2,feynman,nielsen_chuang}.  Quantum computers hold significant promise as they can have polynomial and exponential speed-ups on some problems \cite{shor_algorithm, Grover}. This increase in processing power could have a fundamental impact on computer vision.
 
Small-scale, commercial quantum computers are becoming increasingly available. The challenge for the computer vision community is to identify or create algorithms that are well suited to quantum computing and can exploit potential benefits. Unfortunately, this is non-trivial for three reasons. Firstly, quantum computers use qubits instead of bits, which is a fundamentally different way of storing and processing data. Furthermore, existing quantum computers and quantum emulators can only process a very small amount of data, making it challenging to design and test algorithms. Finally, current and near term quantum computers will be noisy and algorithms should be robust to noise.
 
This paper proposes a novel method for image segmentation that is a natural fit for quantum computation, is robust to noise and can be run on current and near-term quantum computers. The graph cut image segmentation methods of max-flow min-cut and normalized cuts are mapped to the Quantum Approximate Optimization Algorithm (QAOA) \cite{Farhi2014}. QAOA is an attractive choice for its resilience to systematic noise \cite{Farhi2017_hardware} and its potential to outperform classical algorithms \cite{Farhi2016}. The scalability of QAOA is explored in \cite{performance}, which suggests that it is indeed scalable and therefore, in principle, suitable for running segmentations on larger images. The paper demonstrates a practical application for quantum algorithms in the field of computer vision and provides image segmentation results on synthetic and medical images. The methods work on current computers and can scale as larger computers become available. 
 
Quantum computing will open up new research areas in computer vision and lead to the revisiting of existing techniques. This paper is a first step towards ever more sophisticated computer vision algorithms for the new promising quantum hardware.

\section{Related Work}
Performing computer vision tasks on a quantum device is a novel concept. To our knowledge, this is the first implementation of image segmentation with a quantum algorithm. Nevertheless, there have been previous implementations on other image recognition tasks with quantum approaches. Several works \cite{Neven2008, Neven2008_1, Google2009, Grant2018} have implemented image classification with quantum devices, including quantum annealers. These use a different framework to the gate-based devices that are discussed in this work. Image matching has also been investigated with the quantum annealer \cite{Neven2008}. Moreover, quantum-inspired classical approaches have been used in image segmentation \cite{Aytekin2013} and edge-detection \cite{Yuan2013, Fu2009}. These still make use of classical computers, although they take inspiration from quantum concepts.
Finally, a proposed algorithm promises an exponential speed-up for edge detection \cite{Zhang2014}, however requires improvements in quantum hardware beyond what is currently available. This work differs from the latter as it is a near-term approach that is already feasible now on current devices.

Related to QAOA, there has been previous work on graph cut problems, in particular the unweighted MaxCut problem on 3-regular graphs \cite{Farhi2014}. This gave cuts that were less optimal than the best classical algorithm. However, the results are expected to improve with larger $p$ (see Section \ref{section:QAOA}), which the author has left unexplored.
Other theoretical work includes the contributions by Wang et al. \cite{fermionic_view}, who have obtained analytical expressions for the QAOA objective function for several specific cases. 
Weighted MaxCut problems have also been investigated with QAOA prior to this work for finding clusters in data \cite{Otterbach2017}. 

\section{Preliminaries}
\subsection{Introduction to Quantum Computation}
Current algorithms for classical computers involve high-level instructions, as the field has matured to the stage that programming can be performed without considering the operations of gates and circuits. However, the situation is different for quantum computers, as the infrastructure for this luxury has not yet been set in place. Therefore, algorithms described for quantum computers still include concepts such as bits, gates and circuits.

Within the quantum computation paradigm, qubits replace bits as the building blocks of information. Within this framework, qubits are denoted ``states".   Analogously to the classical bits, the quantum bits $\ket{0}$ and $\ket{1}$ exist. These can respectively be written in vector notation as the following: $\begin{bmatrix}
1 \\
0
\end{bmatrix}$
and
$\begin{bmatrix}
0 \\
1
\end{bmatrix}$
. These vectors form a basis in $\mathbb{C}^2$, and this is commonly referred to as the computational basis. 

Quantum states cannot be known unless measured. This is analogous to reading the bits in classical computation. Prior to a measurement, the state $\ket{\psi}$ would be in a superposition, which is a linear combination of the $\ket{0}$ and $\ket{1}$ states. More precisely, a state in superposition is given by
$\ket{\psi} = a_1 \ket{0} + a_2 \ket{1}$, where $a_1, a_2 \in \mathbb{C}$. Using the vector notation for $\ket{0}$ and $\ket{1}$, it can be expressed as a vector $\ket{\psi} = \begin{bmatrix}
a_1 \\
a_2
\end{bmatrix}$. 
The physical meaning of this is that the qubit, when measured, will yield ``0'' with probability $|a_1|^2$ and ``1'' with probability $|a_2|^2$. For normalization purposes, $|a_1|^2 + |a_2|^2 = 1$. A curious feature of quantum mechanics is that measurement inherently changes and destroys the state, so that a result of ``0'' transforms the initial state $\ket{\psi}$ to become $\ket{0}$.

In quantum computation, states are ``evolved'' to other states using gates. A gate is therefore a mapping from one state to another, and can be expressed as a unitary matrix. Universal sets of gates exist, so that any gate can be decomposed into a combination of elements from a given set. These sets are analogous to the \textsc{nand} gate in classical computation. 

Useful gates for our discussion are the Pauli $\sigma^x$, $\sigma^y$ and $\sigma^z$ gates:
$\sigma^x =
\begin{bmatrix}
 0 & 1 \\
 1 & 0 \\
\end{bmatrix}
$, $\sigma^y =
\begin{bmatrix}
 0 & -i \\
 i & 0 \\
\end{bmatrix}
$,
$\sigma^z = 
\begin{bmatrix}
1 & 0 \\
0 & -1\\
\end{bmatrix}$. Also relevant is the Hadamard gate, given by $\frac{1}{\sqrt{2}}
\begin{bmatrix}
 1 & 1 \\
 1 & -1 
\end{bmatrix}$
.

%The $\sigma^x$ operator makes the transformation $\ket{0} \leftrightarrow \ket{1}$,
%whereas the $\sigma^z$ imparts a negative sign on the $\ket{1}$ state, whilst leaving the $\ket{0}$ state unchanged. Together with the identity operator, these operators form a basis for the $2\times2$ complex matrices.
Another important class of operators is that of Hamiltonians $\hat{H}$. These have real eigenvalues that correspond to the energies of its eigenstates.

The notation $\bra{\phi}$ indicates the Hermitian conjugate of the state $\ket{\phi} = \begin{bmatrix}
 b_1 \\
 b_2
\end{bmatrix}$, so that $\bra{\phi} = \begin{bmatrix}
 b_1^* \hspace{0.3em}
 b_2^*
\end{bmatrix}$ for $b_1,b_2 \in \mathbb{C}$ . The visually similar $\ket{\psi}\bra{\phi}$, on the other hand, is the outer product of the vectors, giving a matrix.
For a more rigorous and detailed introduction to quantum computation, please see \cite{nielsen_chuang}.

\subsection{Outline of QAOA}
\label{section:QAOA}
%With QAOA, the problem we attempt to solve is encoded in a matrix $\hat{H}_z$. There is also a matrix $\hat{H}_x$, which determines the initial state of the qubits. This is often a standard one. Gates are applied to the initial state, according to these matrices. The gates are parametrised by the angles $\{\bm{\gamma},\bm{\beta}\}$, which can be optimised with a classical, non-quantum computer. The resulting output qubits then encode the solution to the problem. 

QAOA is an algorithm that solves problems in combinatorial optimisation. In particular, it aims to maximise objective functions that map a bitstring $z = z_1...z_n$ to $\mathbb{R}$. With this in mind, the objective function $C(z)$ is encoded as a diagonal $2^n \times 2^n$ matrix $\hat{H}_z$, with each diagonal element representing the value of the objective function evaluated at a bitstring. For instance, in the setting of two qubits, $\hat{H}_z$ becomes $\hat{H}_z = diag([C_{00}, C_{01}, C_{10}, C_{11}])$, where $diag(\cdot)$ constructs a diagonal matrix given the vector of diagonal entries.

By design, this matrix has eigenvectors given by $\ket{z}$ with corresponding eigenvalues $C(z)$. 
Maximizing the classical objective function then corresponds to finding the eigenstate $\ket{z*}$ that gives the largest eigenvalue.
$\hat{H}_z$ is a Hermitian matrix, and this also yields the desired property that all eigenvalues $C(z)$ are real. It is often denoted the cost Hamiltonian.

The strategy of QAOA is to start with the highest eigenstate $\ket{\psi_i}$ of another matrix $\hat{H}_x$, which should be easy to construct. This state is then appropriately evolved to the highest eigenstate of $\hat{H}_z$. 

The matrix $\hat{H}_x$ is often a standard matrix independent of the problem, given by $\sum_{j=1}^n \sigma_j^x$, with highest eigenstate $\ket{\psi_i} = \left[ \frac{\ket{0}+ \ket{1}}{2} \right]^{\otimes n}$. The superscript denotes an $n$-fold tensor product. This state can be easily constructed through the application of the Hadamard gate on each qubit. $\hat{H}_x$ is also made to be a Hermitian matrix with real eigenvalues, and often referred to as the driver Hamiltonian.

As the real evolution between the two states is unknown, we apply an approximate unitary $U(\bm{\gamma},\bm{\beta})$ that asymptotically converges to the real unitary operator. Here, $\bm{\gamma}$ and $\bm{\beta}$ are sets of angles defined by $\bm{\gamma} = \{\gamma_1,..,\gamma_p\}$ and $\bm{\beta} = \{\beta_1,..,\beta_p\}$. Making use of this approximate unitary means that one can only expect to obtain a final state  $\ket{\bm{\gamma}, \bm{\beta}}$ that is similar, but not identical to the actual eigenstate $\ket{z*}$. In short, 
\begin{equation}
\ket{\bm{\gamma}, \bm{\beta}} = U(\bm{\gamma}, \bm{\beta})
\ket{\psi_{i}}.
\end{equation}
%The standard $\hat{H}_x$ matrix is the sum of one qubit Pauli $X$ matrices, $\hat{H}_x = \sum_{j=1}^n \sigma_j^x$.
%The eigenstate with the largest eigenvalue is simply an equally weighted superposition of the basis states: $\ket{\psi_i} = \frac{1}{\sqrt{2^n}}\sum_z \ket{z}$.  This can be easily constructed with the Hadamard gate applied on each qubit. 
%The Hadamard gate has the matrix representation:
%$\frac{1}{\sqrt{2}}
%\begin{bmatrix}
% 1 & 1 \\
% 1 & -1 
%\end{bmatrix}
%.
%$
The exact expression that we use for the approximate unitary is as follows:
\begin{equation}
    U(\bm{\gamma}, \bm{\beta}) = e^{-i\gamma_p \hat{H}_x} e^{-i\beta_p \hat{H}_z} ... e^{-i\gamma_1 \hat{H}_x} e^{-i\beta_1 \hat{H}_z}
    \label{eq:QAOA_evolution}.
\end{equation}
%Indeed, the angles in $\bm{\gamma}$ are periodic in $\pi$ and the angles in $\bm{\beta}$ have a period of $2 \pi$. %In the implementation, these angles are initialized uniformly at random within the first period.

\begin{figure}[h]
    \centering
 	\begin{tikzpicture}[->,thick, auto]
\node [out-block] (input) {Input: $(\bm{\gamma_1, \beta_1})$};
\node [back, below=0.6cm of input.south] (back1) {};
\node [below=of back1.north, xshift=-2.9cm,yshift=1.0cm] {for $i$ in \{1, ..., $N$\}};
\node [block, below=of back1.north, xshift=0cm, yshift=0.6cm] (quant1) {Quantum computer: evaluate $F_p(\bm{\gamma}, \bm{\beta})$};
\node [block, below=0.8cm of quant1] (class1) {Classical computer: maximization step of $F_p(\bm{\gamma}, \bm{\beta})$ wrt ($\bm{\gamma}, \bm{\beta}$)};

\node [back2, below=1cm of back1] (back2) {};
\node [below=of back2.north, xshift=-2.9cm,yshift=1.0cm] {for $j$ in \{1, ..., $M$\}};
\node [block, below=of back2.north, xshift=0cm, yshift=0.6cm] (quant2) {Quantum computer: construct $\ket{\gamma_{N+1}, \beta_{N+1}}$};
\node [block, below=0.8cm of quant2] (meas) {Measure bitstring};
\node [out-block, below=0.6cm of back2.south] (output) {Output: most frequently measured bitstring};

\path[every node/.style={font=\sffamily\small}]
    (input.south) edge node [below] {} (back1.north)
    (quant1.west) edge[bend right] node [left] {$F_p(\bm{\gamma}_{i}, \bm{\beta}_{i})$} (class1.west)
    (class1.east) edge[bend right] node [right] {$(\bm{\gamma}_{i+1}, \bm{\beta}_{i+1})$} (quant1.east)
    (back1.south) edge node [right] {$(\bm{\gamma}_{N+1},\bm{\beta}_{N+1})$} (back2.north)
    (quant2.south) edge node [right] {$\ket{\gamma_{N+1},\beta_{N+1}}$} (meas.north)
    (back2.south) edge node [below] {} (output.north);
%\path[line] (quant1.south) edge[bend right] node [below] {$F_p(\gamma_{i+1}, \beta_{i+1})$} (class1.south);
%\path[line] 
%\draw[->] (quantum1.south) -- (classical1.south);
\end{tikzpicture}
    \caption{Flowchart of the QAOA algorithm. The upper box outlines the hybrid classical-quantum approach, where the quantum computer evaluates the objective function and the classical computer optimizes it in a step-wise manner for a fixed number of steps  $N$. Given the optimized parameters, the quantum computer then constructs the corresponding state and measures it, returning a bitstring. This procedure is repeated a fixed number of steps $M$ and the most frequently measured bitstring is the output of the algorithm.}
    \label{fig:qaoa_outline}
\end{figure}
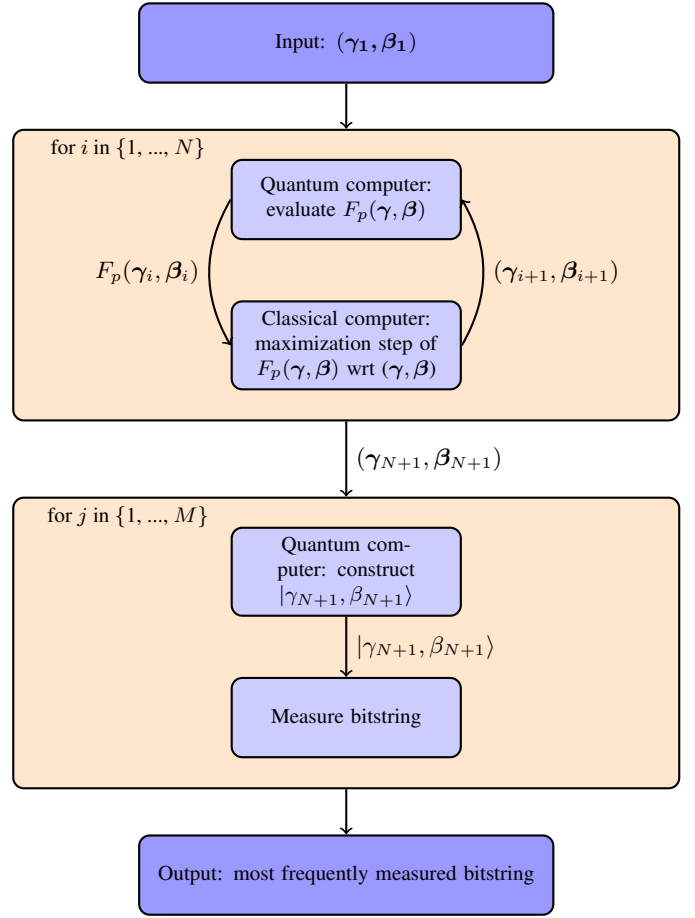

The objective function is then defined as:
\begin{equation}
    F_p(\bm{\gamma}, \bm{\beta}) = \bra{\bm{\gamma}, \bm{\beta}} \hat{H}_z \ket{\bm{\gamma}, \bm{\beta}}.
\end{equation}
This is optimized with respect to the angles $\bm{\gamma}$ and $\bm{\beta}$, to find $\ket{\bm{\gamma}, \bm{\beta}}$. Since this state should have a large overlap with $\ket{z*}$, measuring this state multiple times allows us to obtain the most frequent bitstring, that is hopefully equal to the optimal bitstring. The bitstring encodes the segmentation result, with the vertices corresponding to the background having a value of 0 and those representing the object having a value of 1. Fig. \ref{fig:qaoa_outline} shows a flowchart of the algorithm, with the two main subroutines of the algorithm displayed in the two boxes.

\begin{algorithm}
\begin{algorithmic}
\STATE Input: $p, \hat{H}_x, \hat{H}_z$.
\STATE Initialize $\boldsymbol{\gamma}_0$ and $\boldsymbol{\beta}_0$ randomly.
\FOR{$i \in \{0,...,N_1\}$, }
\STATE Construct objective function $F_p(\boldsymbol{\gamma_i}, \boldsymbol{\beta}_i)$ with quantum computer.
\STATE Maximization step of $F_p(\boldsymbol{\gamma_i}, \boldsymbol{\beta}_i)$ to give $(\boldsymbol{\gamma}_{i+1}, \boldsymbol{\beta}_{i+1})$
\ENDFOR
\FOR{$j \in \{0,...,N_2\}$,}
\STATE Make state $\ket{\boldsymbol{\gamma}, \boldsymbol{\beta}}$ with quantum computer.
\STATE Measure in computational basis to obtain bitstring.
\ENDFOR
\STATE Output: most frequently measured bitstring.
\end{algorithmic}
\caption{QAOA}
\end{algorithm}

Considering $M_p = \max_{\bm{\gamma}, \bm{\beta}} F_p (\bm{\gamma},\bm{\beta})$, we can see that $M_{p-1} \leq M_p$, since the former can be obtained from the latter by setting the first two angles in (\ref{eq:QAOA_evolution}) to zero. Therefore, the approximation can only improve with increasing $p$. 

\subsection{Graph Cut Methods}
For max-flow min-cut \cite{BoykovJolly2001}, the image is represented as a graph $\mathcal{G}=(V,E)$ where $V$ denotes the set of vertices and $E$ the undirected edges. Each edge $e \in E$ is a tuple $e = (a, b, w_e)$, where $a$ and $b$ denote the vertices that the edge connects and $w_e$ is the weight of the edge. Equivalently, we will make use of the notation $w(a,b)$ to refer to this weight. A cut $C(A,B)$ is a subset of the edges $E$, such that the terminal vertices $a \in A$ and $b \in B$ are in two disjoint subsets. After having constructed the graph, the minimum cut is taken, which corresponds to the cut that severs the minimal weight edges: $\mathrm{min}_{C \subset E} (\sum_{e\in C} w_e)$.

To convert the image to its graph representation, each pixel in the image is first represented as a vertex in the graph. These vertices will be denoted $P$. We consider only the 4-neighbourhood edges for each pixel, and denote these edges as ``$n-$links''.
The source and sink vertices, also termed the terminal vertices, are also defined, and these are vertices that represent the background and object respectively. We form edges between all pixel vertices and each terminal vertex, and denote these as ``$t-$links''. 

The weights for an $n-$link are given by a similarity function in terms of the pixel intensities of vertices $a, b \in P$ : $w(a,b) = \mathrm{exp}\left(-\frac{(I_a - I_b)^2}{2 \sigma^2}\right)$. The weights for the $t-$links are more complex. If the pixel in question has been chosen by the user to belong either to the foreground or background, the weights are made large so it is unlikely to be cut. For the other $t-$links between pixel $a \in P$ and source $s$, the weights are defined to be $w(a,s) = \lambda \mathrm{log}(Pr(I_a|\mathcal{O}))$, where $\mathcal{O}$ is the set of known pixel intensities labelled ``object'', and the $t-$links between sink and pixel are defined similarly. 
Here $\lambda$ is a scalar giving the relative importance of the $t-$links compared to the $n-$links.
In this project, we will obtain these probability distributions from prior knowledge. 

On the other hand, the normalized cuts technique \cite{Malik2000} performs a cut on the graph consisting only of the vertices representing the pixels. That is, it only considers the $n-$links mentioned previously. 
However, the result of a simple mincut tends to favor small clusters \cite{wuleahy}. To counteract this, normalization terms are added in the objective function, thus penalizing the presence of small clusters. 
The final objective function that is minimized, is then expressed as follows:
\begin{equation}
    \mathrm{N_{cut}}(A,B) = C(A,B) \left( \frac{1}{\mathrm{deg}(A)} + \frac{1}{\mathrm{deg}(B)} \right),
\end{equation}  
where $\mathrm{deg}(A) = \sum_{a \in A} w(a,v)$ for all $v \in V$ (the sum of all the edge weights for the vertices in $A$), and $\mathrm{deg}(B)$ is similarly defined.

\section{Solving Graph Cuts with QAOA}
\label{section:GraphCuts_QAOA}

The max-flow min-cut problem involves the hard constraints that the cut $C(A,B)$ must have the source in $A$ and sink in $B$.
To impose these constraints in QAOA, the driver Hamiltonian $\hat{H}_x$ can be tweaked \cite{Hadfield2017, Babej2018}. Therefore, $\hat{H}_x$ is the standard $\sigma^x$ operator applied on each qubit, apart from those that represent the terminal vertices. In addition, the sink qubit is initialized as $\ket{1}$ and the source qubit as $\ket{0}$. This ensures that the sink and source qubits remain in these states throughout the evolution. The other qubits all start in the equal superposition state of $\frac{\ket{0} + \ket{1}}{\sqrt{2}}$. The cost Hamiltonian $\hat{H}_z$ is that of the weighted maxcut, as given by
\begin{equation}
    \hat{H}_z =\frac{1}{2} \sum_{\langle j,k \rangle} w(j,k) \left( I - \sigma^z_j \sigma^z_k \right),
\end{equation}
where $I$ is the identity over all the qubits $I^{\otimes n}$ and the notation $\langle \cdot, \cdot \rangle$ sums over the vertices that are connected by an edge. To convert the maxcut to the mincut problem, it suffices to make the original edge weights negative. 

For the normalized cuts problem, the cost Hamiltonian $\hat{H}_z$ is turned into the following:
\begin{multline}
    \hat{H}_z = \left(\sum_{\langle j,k \rangle} -w(j,k) \hspace{0.2em} \sigma^z_j\sigma^z_k \right) \\ 
    \left( \sum_{i=0}^{2^n-1} \ket{z_i}\bra{z_i} \left(\frac{1}{ \mathrm{deg}(A_i)} + \frac{1}{\mathrm{deg}(B_i)} \right)\right).
\end{multline}
where $A_i$ and $B_i$ are the subsets of the vertices labelled ``object'' and ``background'' in the bitstring $z_i$.
Written in the diagonal form, the normalizing terms in the second bracket are calculated for each bitstring and multiplied by the existing terms formed from the standard mincut formulation in the first bracket. 
The driver Hamiltonian is still the $\sigma^x$ operator applied on all the qubits, with the initial state of $\left[ \frac{\ket{0}+ \ket{1}}{2} \right]^{\otimes n}$.

We make use of two methods to implement QAOA. First, we use Rigetti's built-in method for the algorithm. This is accessible from the package \texttt{grove}. The implementation makes use of the Quantum Virtual Machine (QVM) \cite{QVM}, which is a quantum simulator. It compiles the unitary gates into elementary gates that can be implemented by Rigetti's quantum devices. We perform the optimization with a Bayesian optimizer, which evaluates the objective function according to its probabilistic belief of the function. This follows the approach of \cite{Otterbach2017}.

Secondly, we implement it using the package \texttt{tensorflow}. We used this framework to leverage the GPU parallelization built into the package. 
It also allows us to make use of the built-in gradient-based optimizers, such as \texttt{AdamOptimizer}. This implementation is particularly useful to simulate the normalized cuts method, with which Rigetti's QVM had difficulties. However, with this approach, we could not simulate as many qubits since the GPU provided encountered memory issues. A subtlety is that this implementation outputs the wavefunction, and the most common bitstring is then taken to be the computational basis state with the largest contribution. The QVM implementation, however, samples from this distribution, giving a more similar result to that of a realistic quantum computer. 

A contentious topic is the evaluation of the gradient for the objective function, which justifies the use of the gradient-based optimizer. The works \cite{circuit_learning, gradient_optimisation1} give methods to evaluate the gradient, although it is not fully clear how practical it is to implement these on near-term devices.

To make the simulation more realistic, we also incorporated noise for the Rigetti QVM implementation of the $p=1$ case. This applies a $\sigma^x, \sigma^y$ or $\sigma^z$ gate after each existing gate with a probability of 0.05 each. It is useful to look at these since any complex matrix, and therefore any error $K$ can be decomposed into these Pauli operators : $K = \alpha I + \beta \sigma^x + \gamma \sigma^y + \delta \sigma^z$ for $\alpha, \beta, \gamma, \delta \in \mathbb{C}$. 

\section{Creation of datasets}
For the implementation of image segmentation, each qubit represents a pixel of the image. However, classically simulating $n$ qubits involves matrices with $2^n \times 2^n$ entries. This difficulty is in part what makes quantum computation powerful. However, for the purposes of this paper, this makes it computationally intensive to segment images of even 16 pixels. Therefore, we needed to work with small-scale images, such as $3 \times 3$ and $4 \times 4$ synthetic images, and $2 \times 7$ croppings of larger medical images. 

\subsection{Bars and Stripes}
The synthetic dataset that is used is called Bars and Stripes. This is created by taking binary data of possible combinations of bars and stripes that stretch over the entire image. For the $3 \times 3$ dataset, there are in total 12 images. We then added a uniform noise between 0.0 up to a maximum of 0.2. The $4\times 4$ dataset is also used, which contains 28 images. 
As the images are essentially binary, the terminal probability distributions are such that 
\begin{equation}
    Pr(I_p|\mathcal{B}) = 
    \begin{cases}
    1,  \mathrm{ if } \hspace{0.2em} I_p > 0.5\\
    0,  \mathrm{ if } \hspace{0.2em} I_p < 0.5
    \end{cases}
\end{equation}
and 
\begin{equation}
    Pr(I_p|\mathcal{O}) = 
    \begin{cases}
    1,  \mathrm{ if } \hspace{0.2em} I_p < 0.5\\
    0,  \mathrm{ if } \hspace{0.2em} I_p > 0.5.
    \end{cases}
\end{equation}

\subsection{Medical Images}
\begin{figure}[h]
\centering
    \includegraphics[width=\linewidth, trim={4cm 2cm 3.4cm 2cm},clip ]{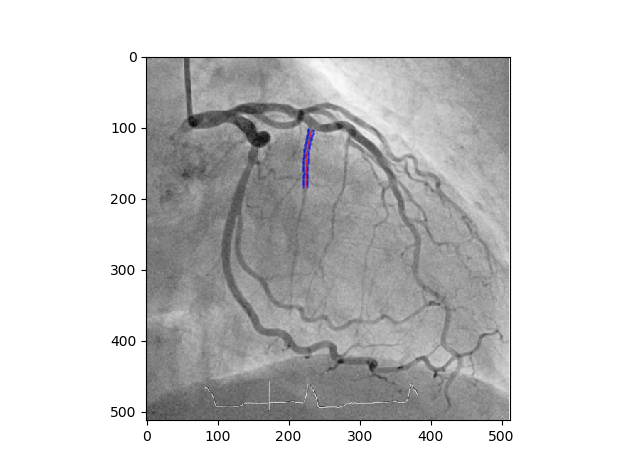}
    \caption{Coronary angiogram with the artery used for segmentation marked in purple.}
    \label{fig:vessel_image_orig}
\end{figure}

The medical images are taken from a coronary angiogram. The image of the artery that we attempted to segment is shown in Fig. \ref{fig:vessel_image_orig}. The midpoints of the artery were found using shortest path algorithms. Then, croppings of the image were made according to the center line of the artery. By segmenting the croppings and combining these results, the segmentation of the entire artery could be reconstructed. 
The cropping consists of splitting it vertically along the midpoint into two sides. The croppings were then taken on each side.
The segmentation was benchmarked against the results found using classical deep learning, which were performed on the entire artery, thus giving the algorithm more context. 

\begin{figure}[h]
    \newlength\figureheight
    \newlength\figurewidth
    \subfloat[Object]{\setlength\figureheight{4.5cm}
    \setlength\figurewidth{4.0cm} % This file was created by matplotlib2tikz v0.7.0.
\begin{tikzpicture}

\definecolor{color0}{rgb}{0.12156862745098,0.466666666666667,0.705882352941177}

\begin{axis}[
height=\figureheight,
tick align=outside,
tick pos=left,
width=\figurewidth,
x grid style={lightgray!92.02614379084967!black},
xlabel={Pixel value},
xmin=0.0, xmax=0.945,
xtick={-0.2,0,0.2,0.4,0.6,0.8,1},
xticklabels={−0.2,0.0,0.2,0.4,0.6,0.8,1.0},
y grid style={lightgray!92.02614379084967!black},
ylabel={\(\displaystyle P(\mathcal{O}|I_p)\)},
ymin=0, ymax=1.0,
ytick={0,0.2,0.4,0.6,0.8,1,1.2},
yticklabels={0.0,0.2,0.4,0.6,0.8,1.0,1.2}
]
\draw[fill=color0,draw opacity=0] (axis cs:0,0) rectangle (axis cs:0.1,0.5);
\draw[fill=color0,draw opacity=0] (axis cs:0.1,0) rectangle (axis cs:0.2,0.615384615384616);
\draw[fill=color0,draw opacity=0] (axis cs:0.2,0) rectangle (axis cs:0.3,1);
\draw[fill=color0,draw opacity=0] (axis cs:0.3,0) rectangle (axis cs:0.4,0.966841186736475);
\draw[fill=color0,draw opacity=0] (axis cs:0.4,0) rectangle (axis cs:0.5,0.568151147098515);
\draw[fill=color0,draw opacity=0] (axis cs:0.5,0) rectangle (axis cs:0.6,0);
\draw[fill=color0,draw opacity=0] (axis cs:0.6,0) rectangle (axis cs:0.7,0);
\draw[fill=color0,draw opacity=0] (axis cs:0.7,0) rectangle (axis cs:0.8,0);
\draw[fill=color0,draw opacity=0] (axis cs:0.8,0) rectangle (axis cs:0.9,0);
\path [draw=black, fill opacity=0] (axis cs:0,0)
--(axis cs:0,1.05);

\path [draw=black, fill opacity=0] (axis cs:1,0)
--(axis cs:1,1.05);

\path [draw=black, fill opacity=0] (axis cs:-0.045,0)
--(axis cs:0.945,0);

\path [draw=black, fill opacity=0] (axis cs:-0.045,1)
--(axis cs:0.945,1);

\end{axis}

\end{tikzpicture}}
    \subfloat[Background]{\setlength\figureheight{4.5cm}
    \setlength\figurewidth{4.0cm} % This file was created by matplotlib2tikz v0.7.0.
\begin{tikzpicture}

\definecolor{color0}{rgb}{0.12156862745098,0.466666666666667,0.705882352941177}

\begin{axis}[
height=\figureheight,
tick align=outside,
tick pos=left,
width=\figurewidth,
x grid style={lightgray!92.02614379084967!black},
xlabel={Pixel value},
xmin=0.0, xmax=0.945,
xtick={-0.2,0,0.2,0.4,0.6,0.8,1},
xticklabels={−0.2,0.0,0.2,0.4,0.6,0.8,1.0},
y grid style={lightgray!92.02614379084967!black},
ylabel={\(\displaystyle P(\mathcal{B}|I_p)\)},
ymin=0, ymax=1.0,
ytick={0,0.2,0.4,0.6,0.8,1,1.2},
yticklabels={0.0,0.2,0.4,0.6,0.8,1.0,1.2}
]
\draw[fill=color0,draw opacity=0] (axis cs:0,0) rectangle (axis cs:0.1,0.5);
\draw[fill=color0,draw opacity=0] (axis cs:0.1,0) rectangle (axis cs:0.2,0.384615384615385);
\draw[fill=color0,draw opacity=0] (axis cs:0.2,0) rectangle (axis cs:0.3,0);
\draw[fill=color0,draw opacity=0] (axis cs:0.3,0) rectangle (axis cs:0.4,0.0331588132635253);
\draw[fill=color0,draw opacity=0] (axis cs:0.4,0) rectangle (axis cs:0.5,0.431848852901485);
\draw[fill=color0,draw opacity=0] (axis cs:0.5,0) rectangle (axis cs:0.6,1);
\draw[fill=color0,draw opacity=0] (axis cs:0.6,0) rectangle (axis cs:0.7,0);
\draw[fill=color0,draw opacity=0] (axis cs:0.7,0) rectangle (axis cs:0.8,0);
\draw[fill=color0,draw opacity=0] (axis cs:0.8,0) rectangle (axis cs:0.9,0);
\path [draw=black, fill opacity=0] (axis cs:0,0)
--(axis cs:0,1.05);

\path [draw=black, fill opacity=0] (axis cs:1,0)
--(axis cs:1,1.05);

\path [draw=black, fill opacity=0] (axis cs:-0.045,0)
--(axis cs:0.945,0);

\path [draw=black, fill opacity=0] (axis cs:-0.045,1)
--(axis cs:0.945,1);

\end{axis}

\end{tikzpicture}}
    \caption{Object and background probability distributions of the medical images, for use in the max-flow min-cut segmentation method.}
    \label{fig:p_fore_back_med}
\end{figure}
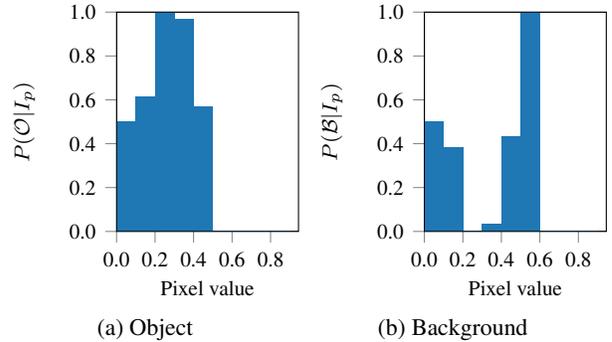

We determined the terminal weights by using the ground truth and the image data from different vessels that were not used for the quantum segmentation task. For the pixels in the object, the intensity values are then binned into a histogram with 10 bins, each with a width of 0.1. This is normalized to turn it into the probability distribution $Pr(I_p|\mathcal{O})$. In this case, the posterior distribution $Pr(\mathcal{O}|I_p)$ is used for the terminal weights, since this gave a better empirical performance. The latter can be obtained from the former through Bayes' theorem. The same procedure is followed for the pixels in the background. 
The terminal probability distributions can be seen in Fig. \ref{fig:p_fore_back_med}.

\section{Results}
\subsection{Bars and Stripes}

\begin{figure}[h]
    \centering
    \subfloat[][Max-flow min-cut]{\includegraphics[scale=0.25]{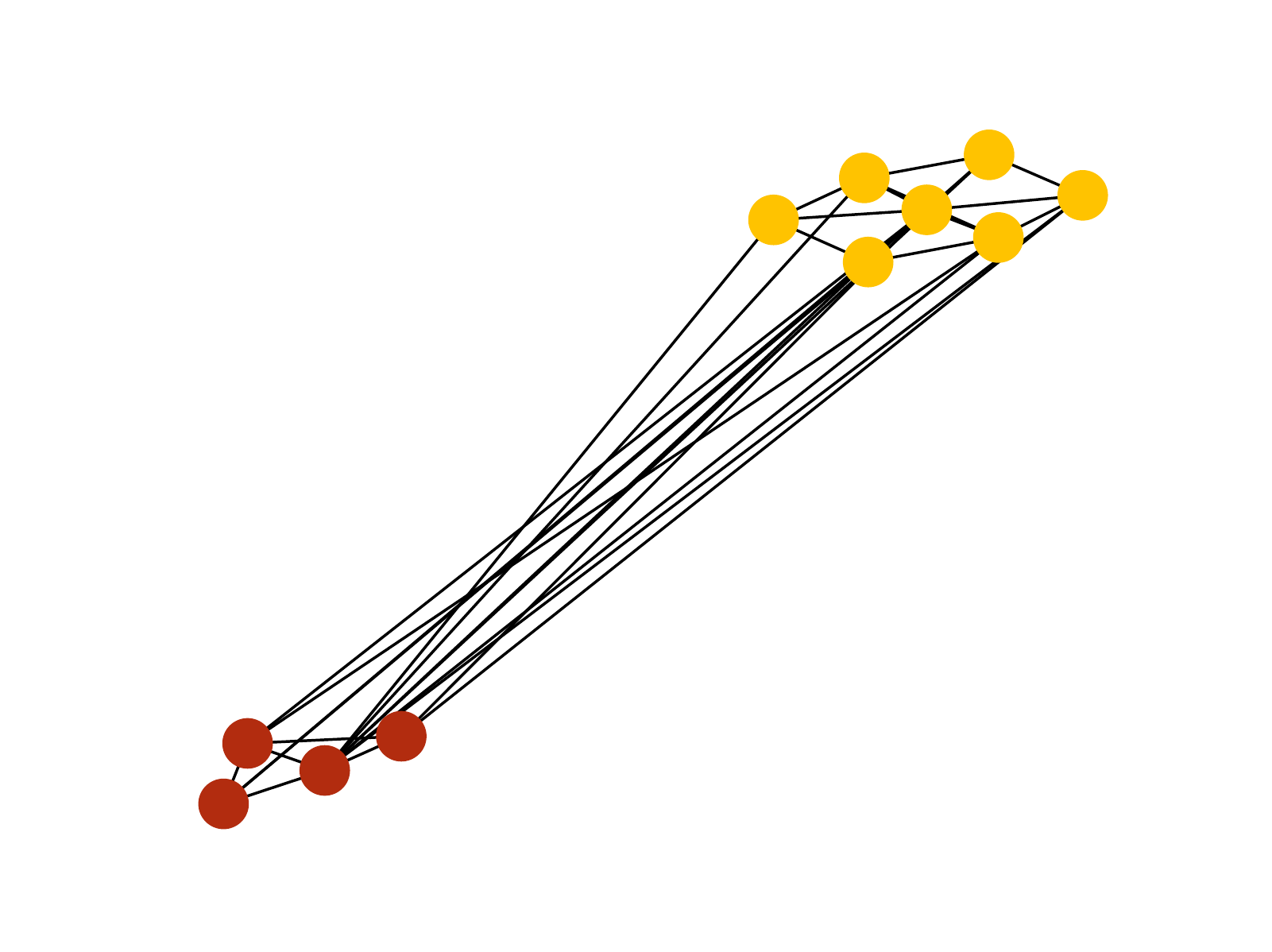}}
     \subfloat[][Normalized cuts]{\includegraphics[scale=0.25]{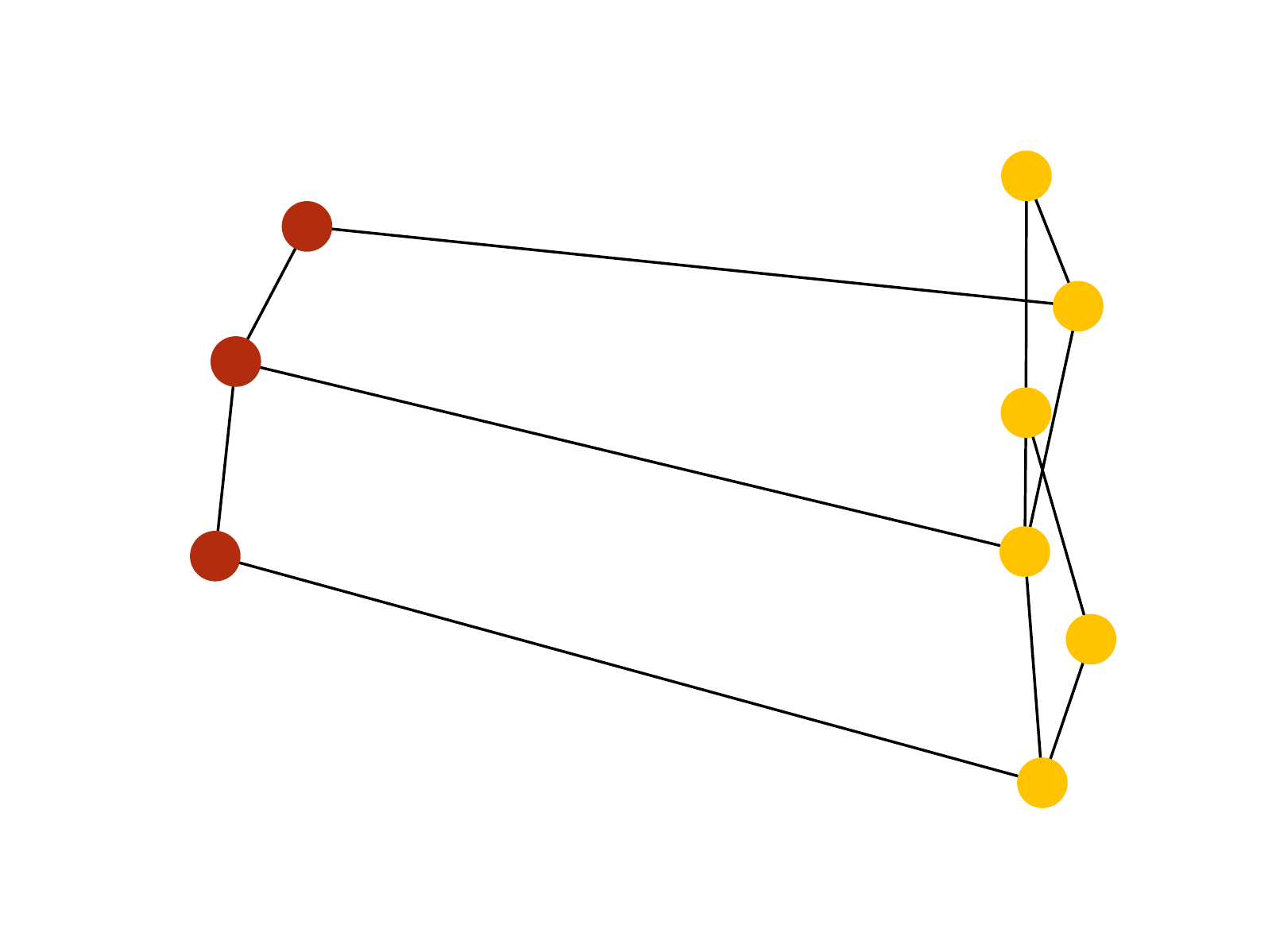}}
    \caption{Graphs for the two approaches. The distances of the vertices are inversely proportional to the weight of their edges. The minimum cut is then taken at the long edges. The two colors represent the two subsets belonging to the object and background.}
    \label{fig:graph_3x3}
\end{figure}

\begin{figure}[h]
    \centering
    \includegraphics[width=\linewidth]{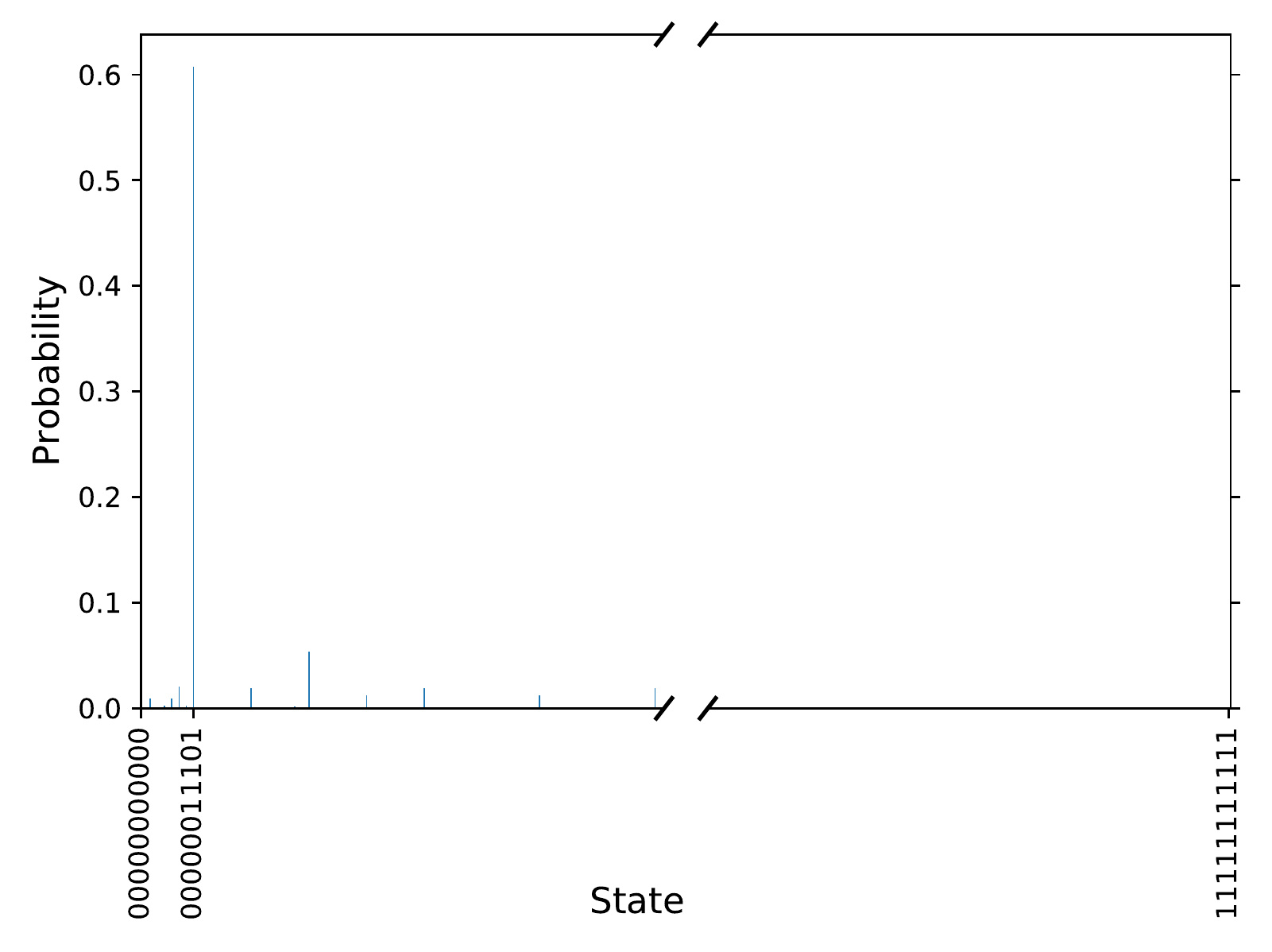}
    \caption{Statistics of all the measured bitstrings for max-flow min-cut. The most frequent bitstring, corresponding to the segmentation result, is represented by the peak. Due to the vast state space, some of the bitstrings are omitted for clarity.}
    \label{fig:stats_3x3}
\end{figure}

\begin{figure}[h]
    \centering
   \includegraphics[width=\linewidth]{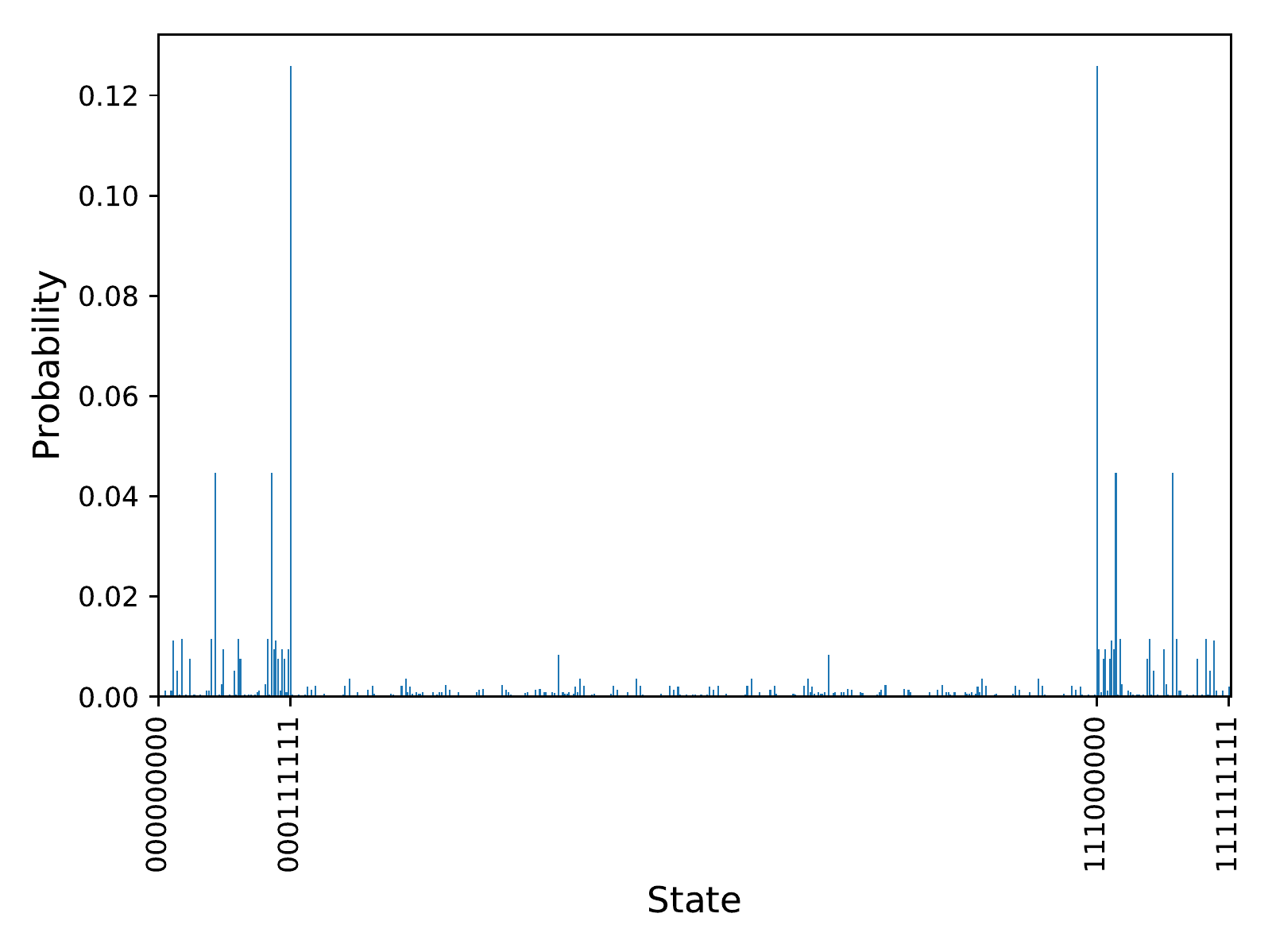}
    \caption{Statistics of all the measured bitstrings for normalized cuts. There are two peaks due to the symmetry of the problem.}
    \label{fig:stats_3x3tf}
\end{figure}

 In Fig. \ref{fig:graph_3x3}, the resulting graphs generated from an example from the dataset can be seen for the two graph cut methods. To show the contribution of the correct bitstring in the overall final output state, Fig. \ref{fig:stats_3x3} shows the probabilities of all the bitstrings in the max-flow min-cut implementation. This shows a peak at the correct bitstring, as well as the vanishing probabilities of the other bitstrings. Note that the last two digits of the correct bitstring are distinct, accounting for the sink and source vertices. 
 Fig. \ref{fig:stats_3x3tf} shows the same histogram for the normalized cuts method. Due to the absence of the terminal vertices, the statistics become fully symmetric, as the method can only find the partitions but cannot assign each partition with the correct label of ``object" or ``background".

The results for the different datasets are shown in Table \ref{tab:final_bars_stripes}. As the optimization results can change each time, we perform an average over multiple runs over all images to calculate the final Dice coefficient. 

For the max-flow min-cut implementation with a Bayesian optimizer, we were able to achieve the same performance as the classical one for all values of steps $p$.
It is interesting to note that even just one step with QAOA can achieve a perfect result. Note that this does not mean that the QAOA final state has a perfect overlap with the ground state. 
The noisy implementation with the probabilistic application of Pauli operators gave a Dice average of 0.72. The $4\times 4$ dataset was again successful, with a Dice average of 1.0.

Optimizing with the gradient-based Adam optimizer was expected to give better results. However, it actually fared worse. 
For one step, a Dice mean of only 0.80 was reached. The results for two and three steps were better, yielding 0.99, which is only marginally worse than the Bayesian optimizer. 

\begin{figure}[h]
    \centering
    \includegraphics[scale=0.12]{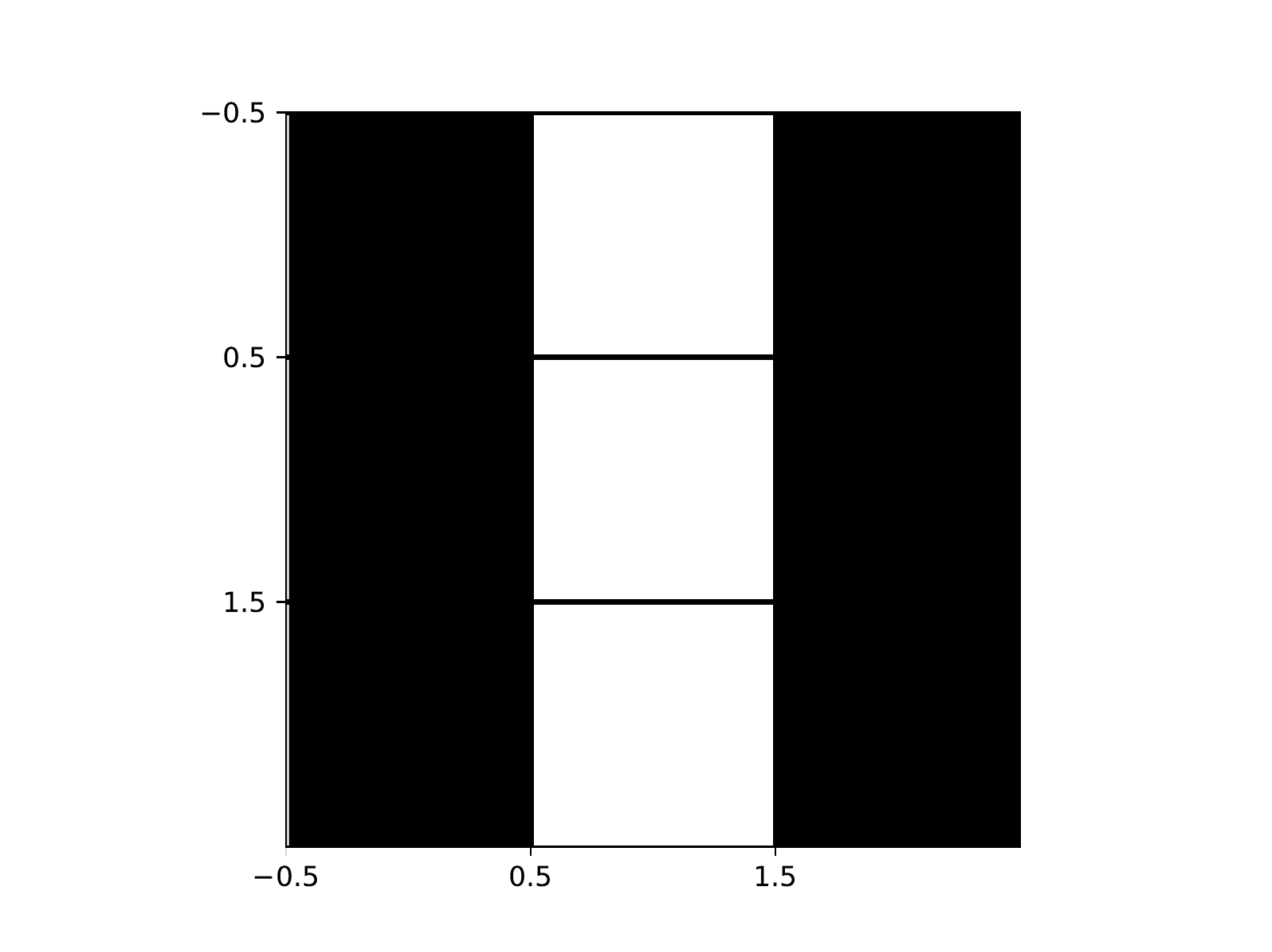}
    \includegraphics[scale=0.12]{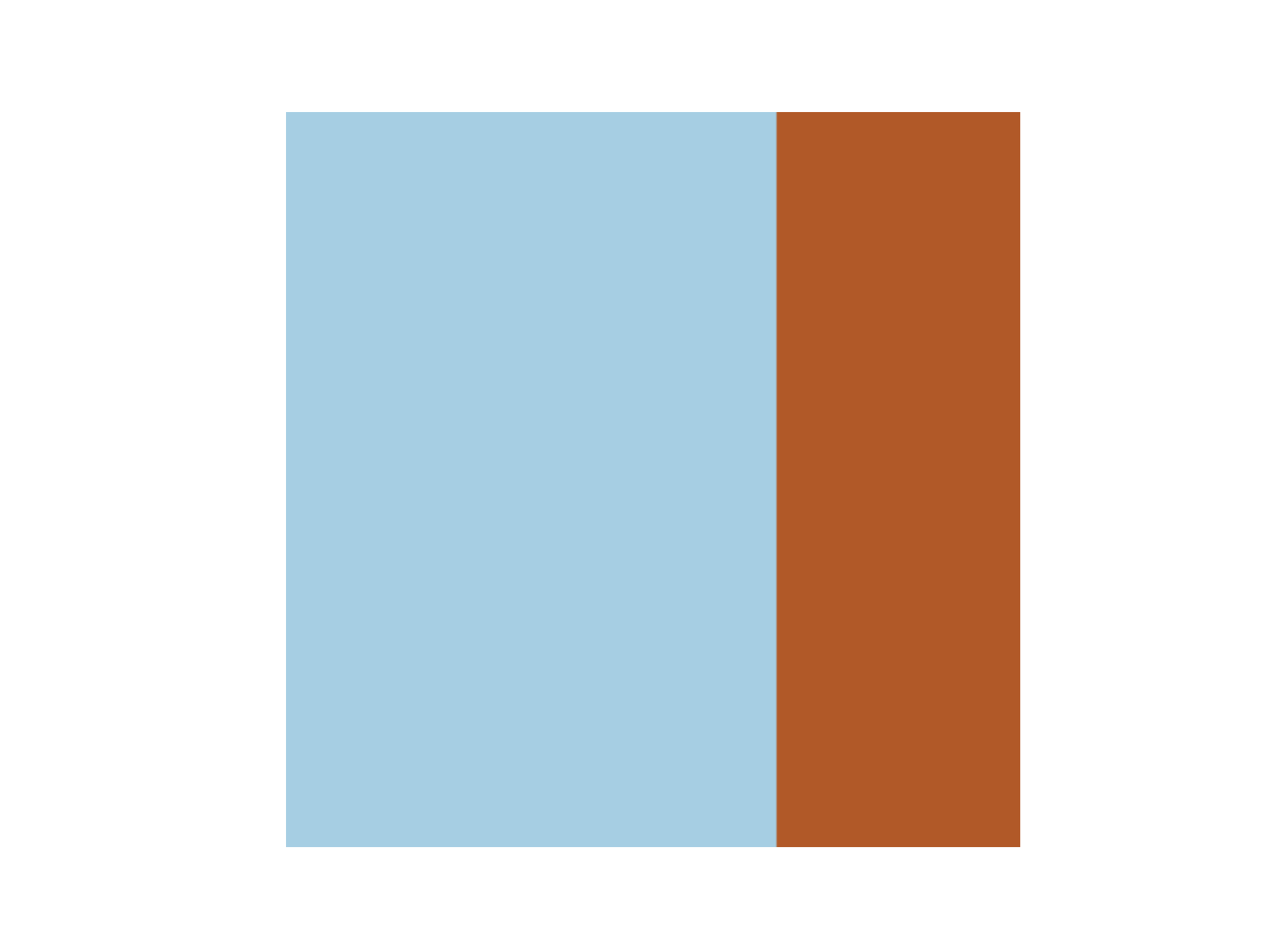}
    \includegraphics[scale=0.12]{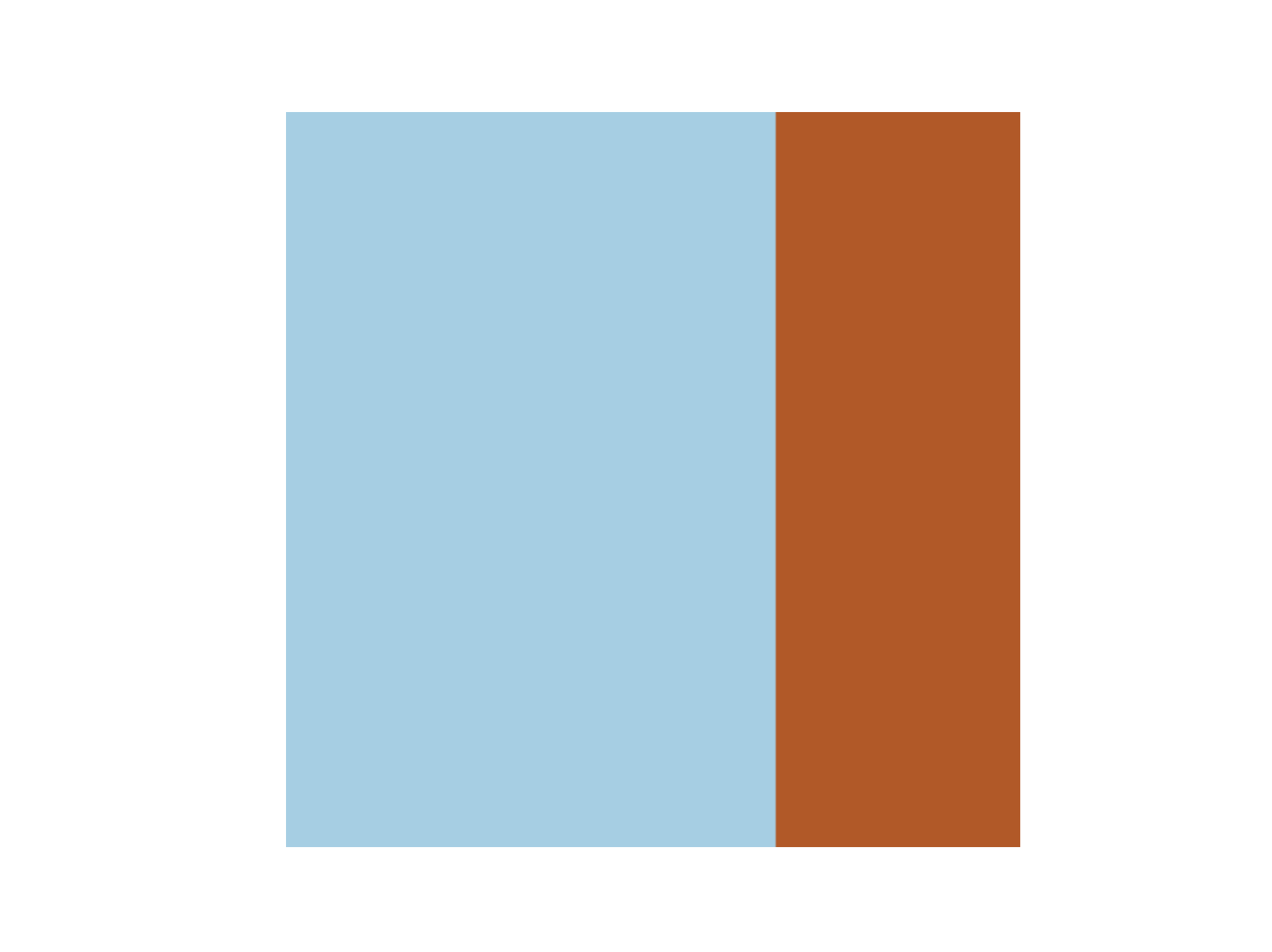}
    \includegraphics[scale=0.12]{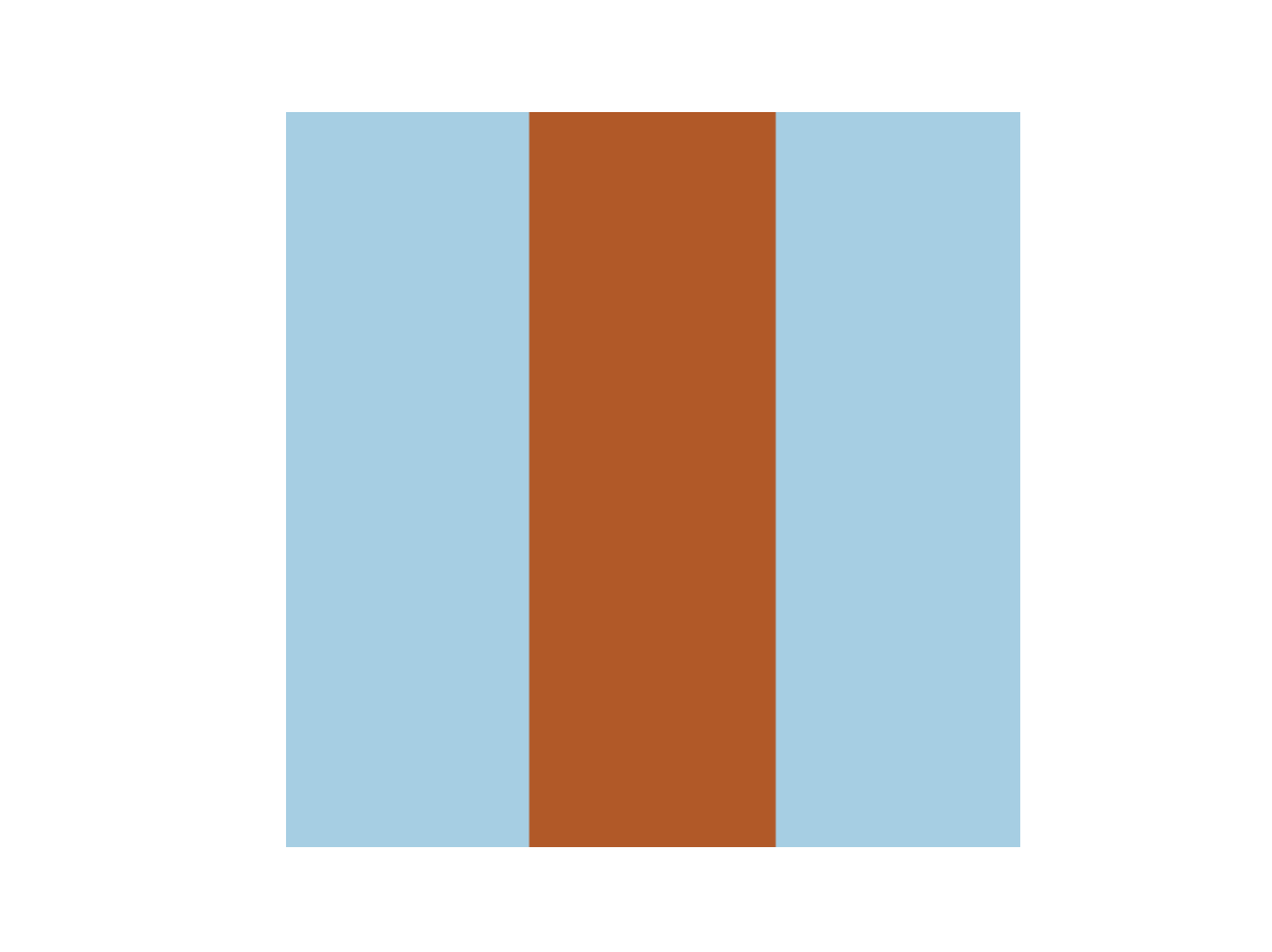}
    \caption{Example of an incorrect quantum segmentation result for the normalized cuts technique, with the results for three runs.}
    \label{fig:normcut_error}
\end{figure}

For the case of the normalized cut, the algorithm could not distinguish between images where there is a stripe or bar in the middle. This was the case for the classical as well as the quantum algorithm. Instead, the segmentation tends to favor stripes to the left or right, as shown in Fig. \ref{fig:normcut_error}. This could possibly be resolved by subdividing the graph further to search for three partitions instead. Also note that the performance of the algorithm seems to have degraded with an increasing number of steps, going from 0.94 to 0.92 and then 0.91, although this is inconclusive due to the large standard deviation. This could be a reflection that the optimization task has increased in difficulty.

\begin{table}[]
    \centering
    \begin{tabular}{@{}l l l l l @{}}
    \toprule Dims & Algorithm & Dice $\mu$ & Dice $\sigma$ & Opt \\ \midrule
     $3\times3$ & maxflow \textit{classical} & 1.0 & 0.0 & None \\ 
     $4\times4$ & maxflow \textit{classical} & 1.0 & 0.0 & None \\ \hline
     $3\times3$ & maxflow QAOA$_1$ & 1.0 & 0.0 & Bayes\\
     $3\times 3$ * & max noisy QAOA$_1$ & 0.72 & 0.15 & Bayes \\
     $3\times3$  & maxflow QAOA$_2$ & 1.0 & 0.0 & Bayes\\
     $3\times3$  & maxflow QAOA$_3$ & 1.0 & 0.0 & Bayes \\
     $4\times4$ * & maxflow QAOA$_1$ & 1.0 & 0.0 & Bayes \\ \hline
     $3\times3$ & maxflow QAOA$_1$ & 0.80 & 0.09 & Adam \\
     $3\times3$  & maxflow QAOA$_2$ & 0.94 & 0.04 & Adam \\
     $3\times3$  & maxflow QAOA$_3$ & 0.99 & 0.01 & Adam \\ 
     \hline
     $3\times3$  & norm cut \textit{classical} & 0.97 & 0.09 & None \\
     $4\times4$  & norm cut \textit{classical} & 0.86 & 0.19 & None \\ \hline
    
     $3\times3$  & norm cut QAOA$_1$ & 0.94 & 0.08 & Adam \\
     $3\times3$  & norm cut QAOA$_2$ & 0.92 & 0.1 & Adam \\ 
     $3\times3$  & norm cut QAOA$_3$ & 0.91 & 0.1 & Adam \\
     
     \bottomrule
    \end{tabular}
    \caption{Result summary for the Bars and Stripes dataset. All statistics are averaged over at least 100 runs for the Adam optimizer. For the Bayesian optimizer, at least 20 runs were performed, except for those marked with *, where it was only possible to perform 3 runs. The mean and standard deviation are taken over the averages of each run. maxflow = max-flow min-cut,  max noisy = max-flow min-cut noisy, QAOA$_p$= QAOA with $p$ steps, norm cut = normalized cuts, Bayes = Bayesian optimizer, Adam = Adam optimizer.}
    \label{tab:final_bars_stripes}
\end{table}

\subsection{Medical Images}
\begin{figure}[h]
     \centering
     \subfloat[][Original cropped image]{\includegraphics[scale=0.25,trim={1cm 4cm 1cm 4cm},clip]{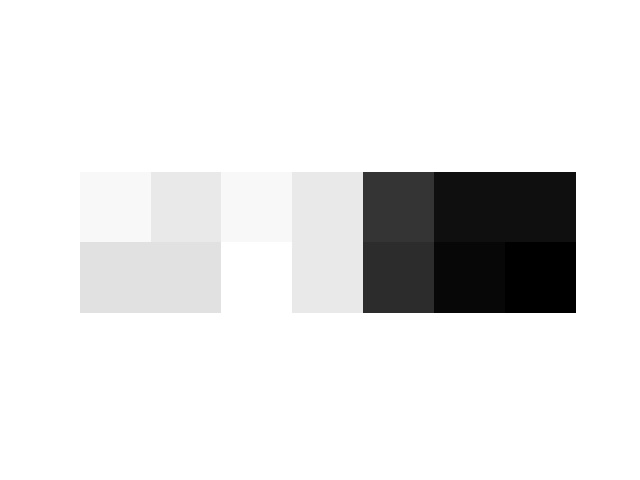}}
      \subfloat[][Classical segmentation]{\includegraphics[scale=0.25, trim={1cm 4cm 1cm 4cm},clip]{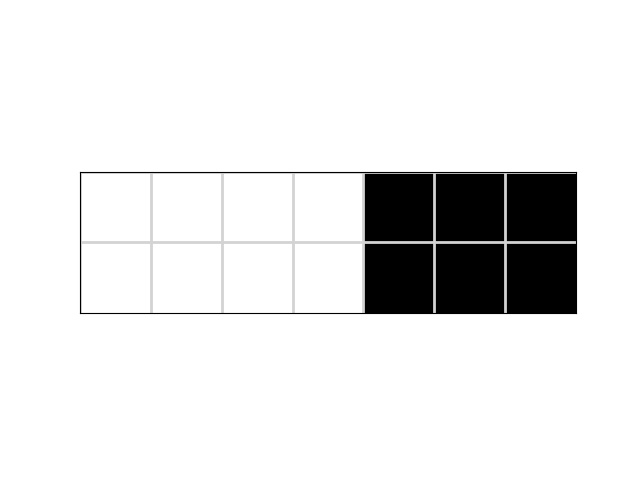}}
      
      \subfloat[][Quantum normalized cuts]{\includegraphics[scale=0.25,trim={1cm 4cm 1cm 4cm},clip]{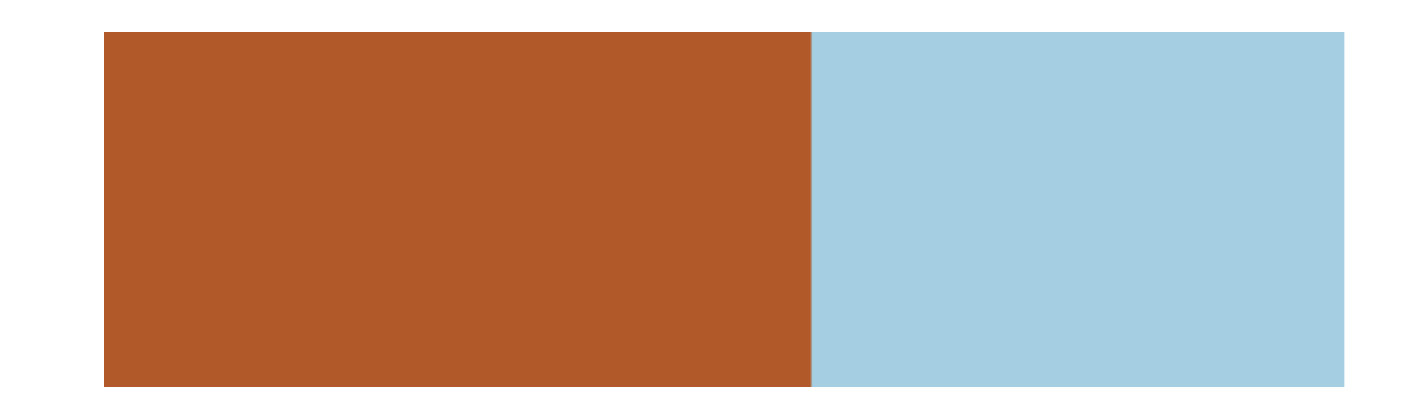}}
      \subfloat[][Quantum max-flow min-cut]{\includegraphics[scale=0.25,trim={1cm 4cm 1cm 4cm},clip]{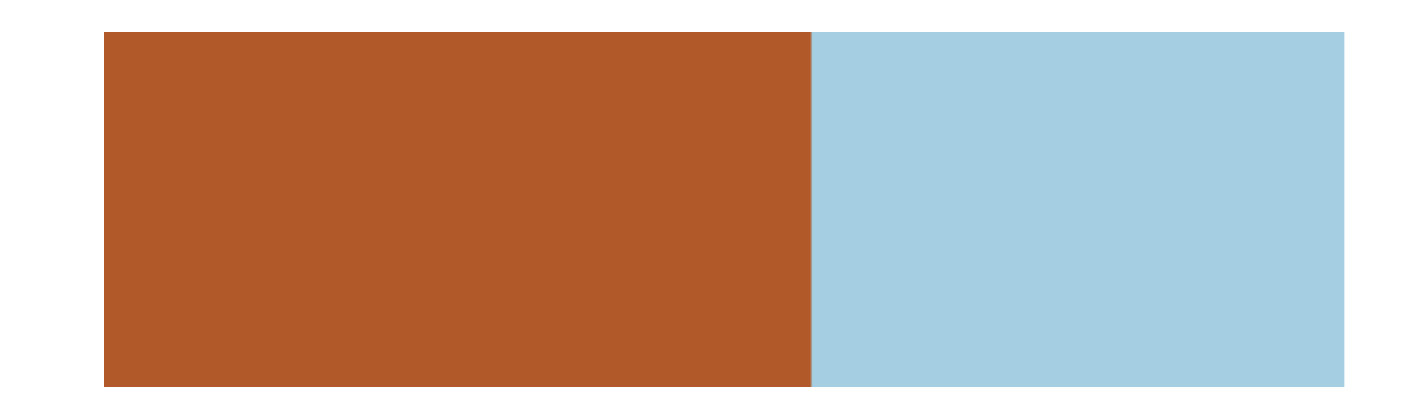}}
     \caption{Successful segmentation result for an example medical image with the normalized cuts approach. This is the cropping to the left hand side of the center line.}
     \label{fig:vessel-1-0}
\end{figure}

\begin{figure}[h]
    \centering
    \subfloat[Max-flow min-cut]{\includegraphics[width=0.5\linewidth,trim={1cm 1cm 1cm 1cm},clip]{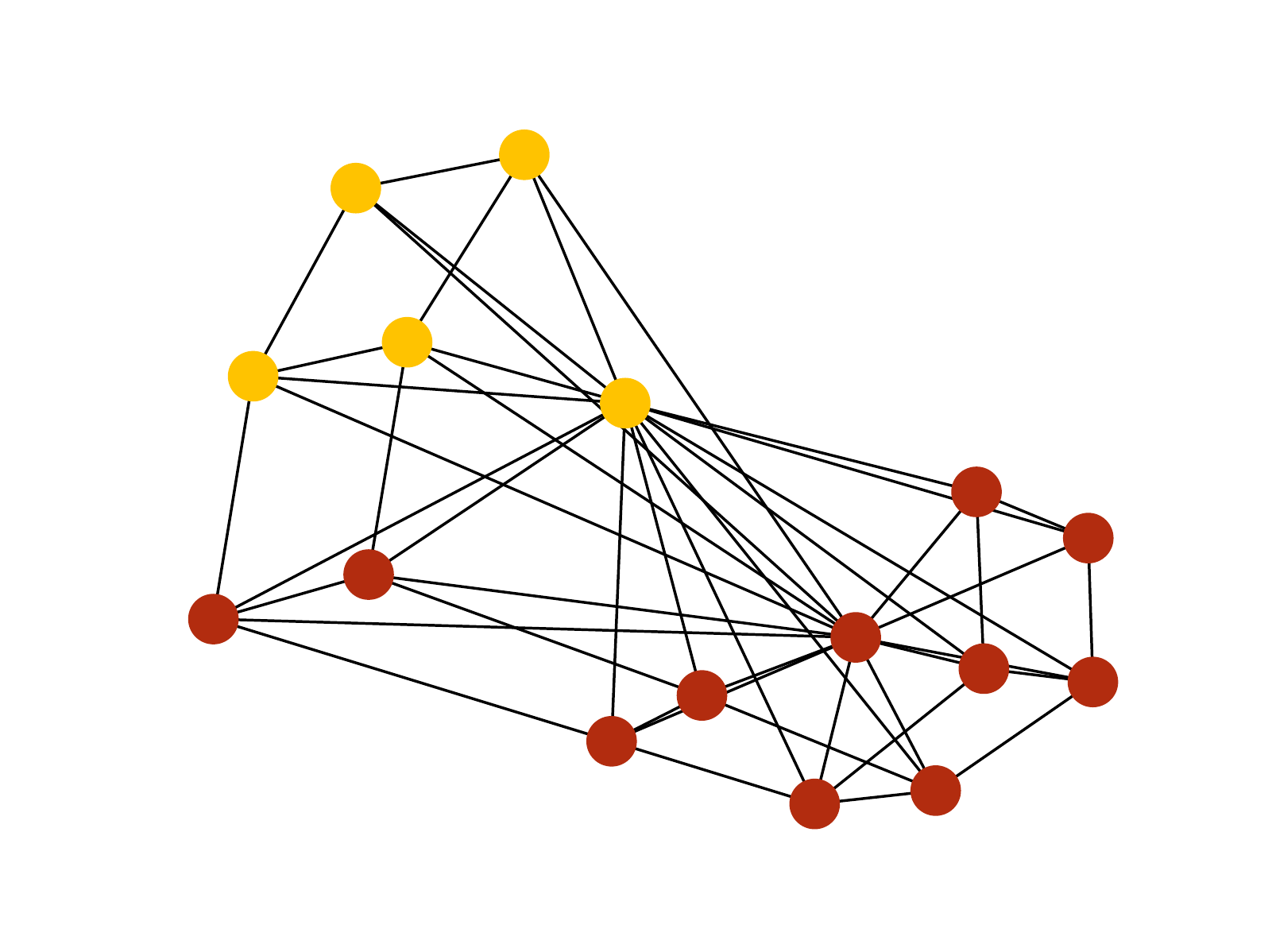}}
    \subfloat[Normalized cut]{\includegraphics[width=0.5\linewidth,trim={1cm 1cm 1cm 1cm},clip]{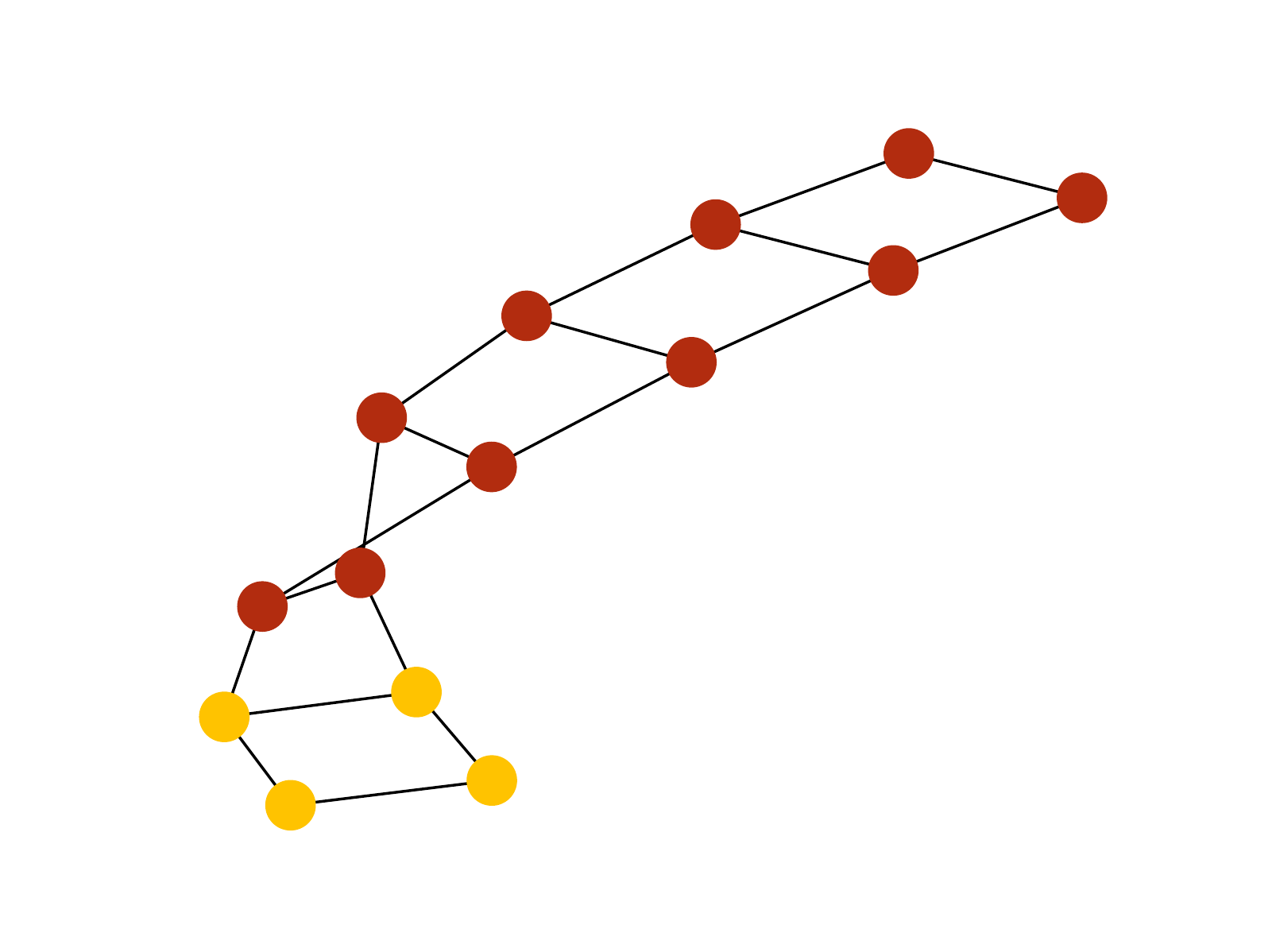}}
    \caption{Underlying graph for one of the images. Again the two different partitions are marked with two distinct colors.}
    \label{fig:graph_vessel1_0}
\end{figure}

For the medical images, both normalized cuts and max-flow min-cut are performed with the Adam optimizer. For both implementations, the results were averaged over three runs. An example cropped image that was segmented is shown in Fig. \ref{fig:vessel-1-0}. The classical, as well as the quantum segmentation results are shown in the same figure. Note that the classical segmentation took the entire artery into account, and therefore has more context.
The underlying graphs are plotted in Fig. \ref{fig:graph_vessel1_0} for the two graph cut methods. This shows that the problem has indeed increased in difficulty compared to the synthetic dataset, owing to the difficulty in identifying clusters within the graph.

\begin{figure}[h]
    \centering
    \includegraphics[width=\linewidth]{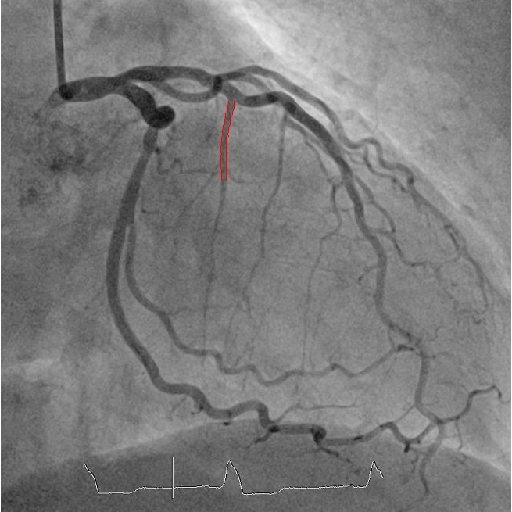}
    \caption{Quantum segmentation result, with the boundaries of the segmentation marked in red on the entire image for the normalized cuts approach.}
    \label{fig:entire_plot}
\end{figure}

\begin{figure}[h]
     \centering
     \subfloat[][Original.]{\includegraphics[width=.45\linewidth]{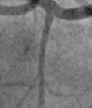}}
     \qquad
     \subfloat[][Classical.]{\includegraphics[width=.45\linewidth]{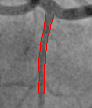}}
 
      \subfloat[][Quantum normalized cuts.]{\includegraphics[width=.45\linewidth]{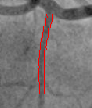}}
      \qquad
      \subfloat[][Quantum max-flow min-cut.]{\includegraphics[width=.45\linewidth]{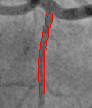}}
     \caption{Comparison of the classical and quantum segmentation results of the artery, with the boundaries of the segmentation marked in red. The result is given for one of the 3 runs.}
     \label{fig:compare}
\end{figure}

The resulting segmentation of the entire artery is shown in Fig. \ref{fig:entire_plot} for normalized cuts. For both implementations, the enlarged segmented image is compared with the classical segmentation result in Fig. \ref{fig:compare}. The Dice coefficient average was 0.88 for both implementations, with a standard deviation of 0.09. A curious feature was that even for the max-flow min-cut implementation, the background and foreground were sometimes confused.
To tackle this ambiguity between the foreground and background, the fact that the right-most (left-most) pixels for the left (right) hand side croppings belonged to the artery was used. 

\section{Discussion}
The segmentation performed with the proposed quantum approach gave encouraging results on artificial and medical data. The Bayesian optimizer has superior performance to the Adam optimizer for the synthetic images. Possible future work includes assessing whether this is due to the nature of the objective function landscape, and whether this is the case for other datasets. The Bayesian optimizer, however, scales less well with more parameters, requiring ever more iterations to find a solution close to the global minimum. On the other hand, easily obtaining the gradient for the Adam optimizer in this setting is an open research question.
 
The max-flow min-cut approach, in general, gave better results than the normalized cuts approach. This was expected as the former makes use of two extra qubits. The probability distribution of the output state given by normalized cuts has in fact two peaks, corresponding to the partitions with the background and foreground labelled interchangeably. In addition, there are a few synthetic images that it cannot segment successfully.
 
Furthermore, one step with QAOA is sufficient on the proposed data, to achieve reliable segmentation results for both the synthetic and medical data. This is encouraging, since fewer steps with QAOA requires fewer gates, and will avoid the increased noise introduced by additional gates.
 
An open and active area of research in the quantum computing community is analyzing the run-time of heuristic quantum algorithms and evaluating their scalability on larger quantum hardware. This is challenging as modelling noise is non-trivial and simulating large quantum computers with classical computers is not computationally tractable. In the context of this paper it would serve to compare its performance with classical problems. Although QAOA was originally developed for NP-hard problems, the max-flow min-cut problem is solvable in polynomial time classically \cite{Elias1956, Ford1956}.  However, the normalized cuts problem is in general NP-complete \cite{wagner_wagner}. Although the quantum approach only offers an approximate solution, it has turned the problem into one of optimizing two parameters for $p = 1$. As the classical algorithm for normalized cuts also looks at approximate solutions, future work can attempt to identify cases where QAOA can outperform the classical algorithm in the quality of the segmentation.
 
The size of image that can be segmented is restricted by the size of quantum computers available today. However, the announcement of larger and more powerful quantum chips from hardware manufacturers such as Google, IBM, Intel and Rigetti is encouraging and will lead to more practical and powerful computer vision algorithms. Such machines will allow better validation of image segmentation on large images and evaluation of the performance of QAOA.
 
\section{Conclusion}
This paper proposes the use of quantum computing for image segmentation, revisiting graph cut algorithms and mapping these to a quantum algorithm that is well suited to noisy near-term quantum devices. The approach is demonstrated on small artificial and medical datasets, where the data size is constrained only by the size of currently available quantum hardware. This paper takes a step towards the practical use of quantum computing in computer vision. As larger scale quantum computers become available, this nascent research topic is expected to grow and lead to significant change in algorithm development.

\section{Acknowledgements}
This work was supported by the Engineering and Physical Sciences Research Council [grant number
868 EP/L015242/1]. Concepts and information presented are based on research and are not commercially available. Due to regulatory reasons, the future availability cannot be guaranteed.

{\small
\bibliographystyle{unsrt}
\bibliography{egbib}
}

\end{document}